\documentclass[letterpaper,10pt,conference]{ieeeconf}
\IEEEoverridecommandlockouts
\usepackage[T1]{fontenc}
\usepackage{float}
\usepackage{amsfonts}
\usepackage{amsthm}
\usepackage{pifont}
\usepackage{xcolor}
\usepackage{soul}

\newtheorem{definition}{Definition}

\newtheorem{problem}{Problem}
\usepackage{glossaries}
\usepackage{listings}
\usepackage{amsmath}
\usepackage{amssymb}
\usepackage{subfig}
\usepackage{graphicx}
\usepackage{stix}

\usepackage{tikz}

\usepackage{etoolbox}
\usetikzlibrary{shapes,arrows}
\usepackage{verbatim}
\usepackage[hidelinks]{hyperref}

\renewcommand{\baselinestretch}{1.0}

\newcommand{\deleted}[1]{}

\begin{document}
\title{\bf 
Multi-rotor Aerial Vehicles in Physical Interactions: A Survey
}
\author{Jiawei Xu 
\thanks{
    J. Xu 
    is with the Autonomous and Intelligent Robotics Laboratory (AIRLab), Lehigh University, PA, USA. Email:
    jix519@lehigh.edu
    }
}
\maketitle
\begin{abstract}
    Research on Multi-rotor Aerial Vehicles (MAVs) has experienced remarkable advancements over the past two decades, propelling the field forward at an accelerated pace. Through the implementation of motion control and the integration of specialized mechanisms, researchers have unlocked the potential of MAVs to perform a wide range of tasks in diverse scenarios. Notably, the literature has highlighted the distinctive attributes of MAVs that endow them with a competitive edge in physical interaction when compared to other robotic systems. In this survey, we present a categorization of the various types of physical interactions in which MAVs are involved, supported by comprehensive case studies. We examine the approaches employed by researchers to address different challenges using MAVs and their applications, including the development of different types of controllers to handle uncertainties inherent in these interactions. By conducting a thorough analysis of the strengths and limitations associated with different methodologies, as well as engaging in discussions about potential enhancements, this survey aims to illuminate the path for future research focusing on MAVs with high actuation capabilities.
\end{abstract}

\section{Introduction}{
    Since the deployment of the first industrial robot sixty years ago~\cite{nof1999handbook}, engineers and researchers have been constantly looking for ways to use robots in processes that can be automated, with the aim of improving productivity in different industries and freeing humans from repetitive tedious tasks. Meanwhile, inspired by birds and insects, engineers have developed aerial vehicles that also bring new features to the industry. One of the aerial vehicle designs that has exhibited high potential in a wide range of applications is multi-rotor aerial vehicles.
    
    \begin{definition}[Multi-rotor Aerial Vehicles (MAVs)]
    MAVs are motorized aerial vehicles that generate thrust force using their rotors directly to compensate for gravity and control motion.
    \end{definition}
    
    In the rest of this paper, terms such as quadrotors, hexarotors, bicopters, and monocopters represent different designs of MAVs based on the authors' choices specific to the article under discussion.
    
    As shown in Figure~\ref{fig:taxonomyRobots}, MAVs occupy a small but significant niche in the realm of modern robots, possessing unique characteristics. With a high level of maturity in studies on MAVs that have opened many possible paths of future development, we would like to answer the following question in this survey:
    \begin{problem}[What is the purpose of MAVs?]
        What tasks are MAVs most suitable for? What are the mechanisms and characteristics of MAVs that make them advantageous in handling these tasks compared to other types of robots? How can developers highlight these characteristics and improve the mechanisms to enhance the effectiveness of MAVs when performing these tasks?
    \end{problem}
    
    Compared to fixed-station manipulators or ground vehicles, MAVs have access to all three axes of 3D space, but they always need to spend energy maintaining stability in flight. Unlike other types of aircraft that require wings as lifting surfaces, MAVs use propellers to generate thrust force directly to counter gravity, reducing the impact of aerodynamic effects and allowing them to hover in place~\cite{yoon2017computational}. However, this also results in low energy efficiency during operation~\cite{sibilski2002comparative} and challenges in achieving high-speed flight~\cite{8793887}. Researchers often emphasize the simplicity of MAVs in design and control~\cite{1302409,4399042,8251877}, their agility~\cite{kushleyev2013towards}, and their scalability~\cite{preiss2017crazyswarm,saldana2018modquad}. Field applications are also a common topic of study for MAVs, such as air surveillance~\cite{semsch2009autonomous}, map classification~\cite{liu2020distributed}, coverage control~\cite{alvissalim2012swarm}, and aerial construction~\cite{zhou2018unmanned,dastgheibifard2018review}, particularly compared to fixed-wing aircraft~\cite{boon2017comparison,5524453,singhal2018unmanned}.
    
    The mixture of advantages and disadvantages of MAVs over other types of robots makes them more suitable for dealing with tasks that require flexible deployment and the ability to follow trajectories on command while carrying a payload. From the perspective of accentuating the positive and avoiding the negative, we find a common ground for the tasks for which MAVs are most suitable, that is, low-speed and short-term \emph{ physical interaction}. 
    \begin{figure}[t!]
        \centering
        \includegraphics[width=\linewidth]{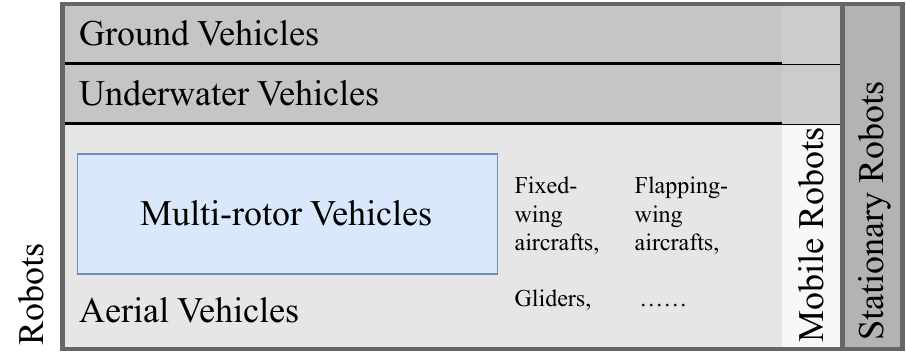}
        \caption{MAVs are but a speck of dust in the modern world of robotic systems, but they open the door to a new field of study in Robotics.}
        \label{fig:taxonomyRobots}
    \end{figure}
    
    One of the most well-explored aspects of physical interaction in the robotics literature is manipulation, especially with fixed-station manipulator arms~\cite{billard2019trends,cui2021toward,kemp2007challenges}. Such robotic arms are capable of achieving robustness and efficiency higher than that of human operators in many scenarios. Meanwhile, research on mobile and aerial manipulations is also increasing both in volume and quality~\cite{holmberg2000development,orsag2017dexterous,hvilshoj2012autonomous}. Physical interaction is a superset of manipulation. \emph{A robot performing manipulation tasks has the intention of changing the states, e.g., the position or orientation, of external objects, while in physical interaction, one may not.} A robot that only changes its own states by creating an internal force/torque without affecting any external object is considered involved in physical interaction but not in manipulation. If a robot makes contact with an object, sliding on the surface of the object without moving it, the robot physically interacts with the object but does not manipulate it. This type of interaction is termed ``contact''~\cite{hogan2022contact,park2008robot}. In scenarios where the states of the object with which a robot is interacting do not matter, such as in a soil/water sampling task, the robot is performing physical interaction with the environment but is not involved in manipulation. Although manipulation constitutes a very important and practical portion of physical interaction, where many open challenges are still under discussion, we will not limit our scope only to manipulation in this survey, but rather to physical interaction involved with MAVs in general. Furthermore, physical interaction includes robot-human interaction~\cite{goodrich2008human,doi:10.1177/0018720816644364,de2008atlas}, robot-environment / object interaction~\cite{vukobratovic2009dynamics,9444288}, and robot-robot interaction~\cite{10014434,8673304}. We will focus on the physical interaction between robots (MAVs in our case) and the physical interaction between MAVs and the environment. Constrained by the two main disadvantages of MAVs, low energy efficiency and low speed operation, the physical interaction that MAVs perform in this study is short-term (typical duration less than $1$ hour) and low-speed (typical speed less than $10$ meters per second).
    \begin{figure}[t]
        \centering
        \includegraphics[width=\linewidth]{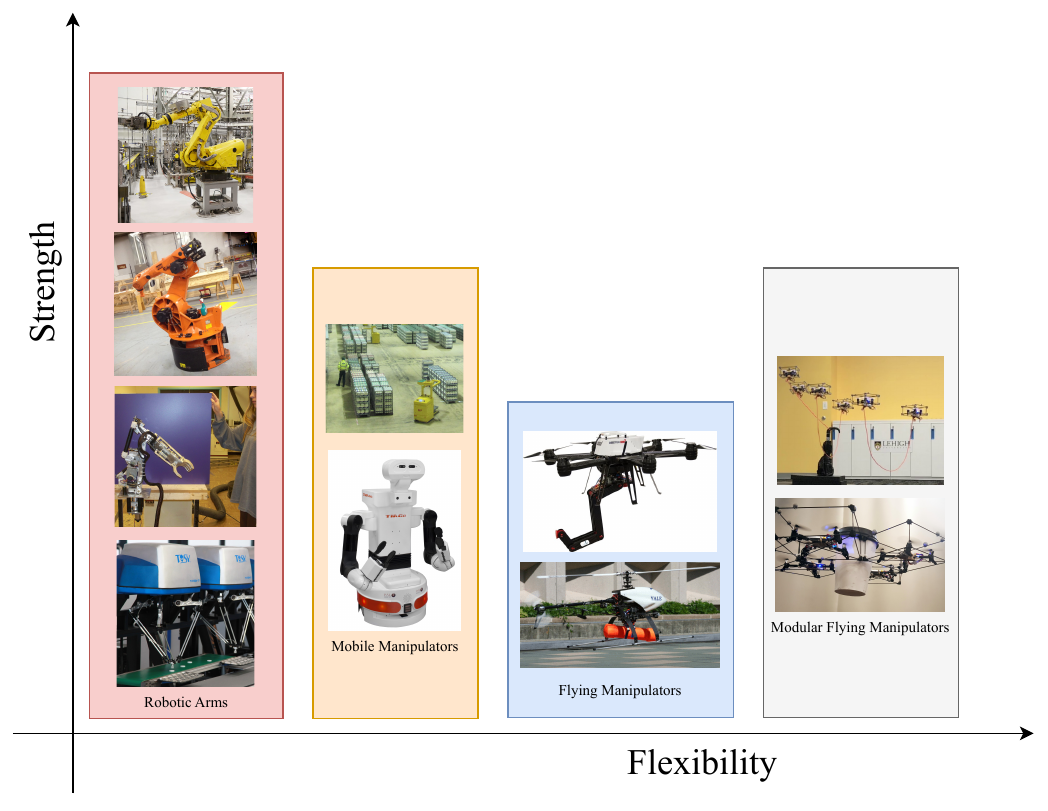}
        \caption{Robots that are capable of physically interacting with objects in the environment. The red, orange, blue, and white colors highlight the decreasing maturity of these solutions.}
        \label{fig:taxonomyPhysicalInteraction}
    \end{figure}
    
    As mentioned above, MAVs possess unique characteristics that allow them to perform physical interactions effectively. As shown in Figure~\ref{fig:taxonomyPhysicalInteraction}, various types of robots are capable of initiating and controlling physical interaction. There always exists a trade-off between the strength, i.e., the payload capacity of the robots, and the flexibility in action, i.e., the mobility of the robots and the range of tasks the robots can tackle. Depending on the application scenario, designers choose specific types of robots for certain tasks. For example, on a production line, speed, strength, and precision are of the highest importance among the robot's requirements. Since the location of the operations and tasks rarely changes, the mobility of the robot is mostly ignored. Therefore, strong stationary industrial manipulation arms and precise delta manipulators dominate the market. Being able to explore the $3$-D space and hover at a position, MAVs are able to carry and operate equipment precisely for physical interaction~\cite{8299552}, barely affected by the change in the location of deployment. 
    Traditional manipulation arms have their maximum strength bottlenecked at each joint, which means that mounting more actuators usually does not increase their operating capacity. In contrast, the addition of actuators to MAVs almost always guarantees increased strength~\cite{segui2014novel} due to the nature of their distributed actuation, resulting in straightforward integration of modularity into MAVs~\cite{ulbrich2011i4copter,saldana2018modquad}, allowing increased MAV strength by adding more modules. In this survey, we will thoroughly examine such characteristics and their corresponding effects on tasks involving physical interaction in different application scenarios through case studies. We will focus on five questions. 
    \begin{enumerate}
        \item What enables the systems to be able to perform physical interaction?
        \item How do we categorize the physical interaction that MAVs perform?
        \item How are MAVs controlled?
        \item What are the special characteristics, if any, of the MAV designs during physical interaction?
        \item What are the possible applications?
    \end{enumerate}
    
    Due to the limit in the scope of research, in some of the cases we study, researchers can unintentionally utilize the special characteristics of MAVs to deal with physical interaction tasks. As a result, usage may be suboptimal. We would like to show how awareness of physical interaction involved in the tasks can possibly further improve these approaches. As the theory of MAVs is becoming more mature, we have seen applications available to consumers. However, we still have to overcome a number of challenges to apply them in a wider range of scenarios where MAVs are theoretically advantageous over other solutions, such as robotic arms. We propose general directions for the development of MAVs in the future with the focus on physical interaction in mind.
}

\section{Physical Interaction}{
    \begin{figure*}[t]
        \centering
        \includegraphics[width=\linewidth]{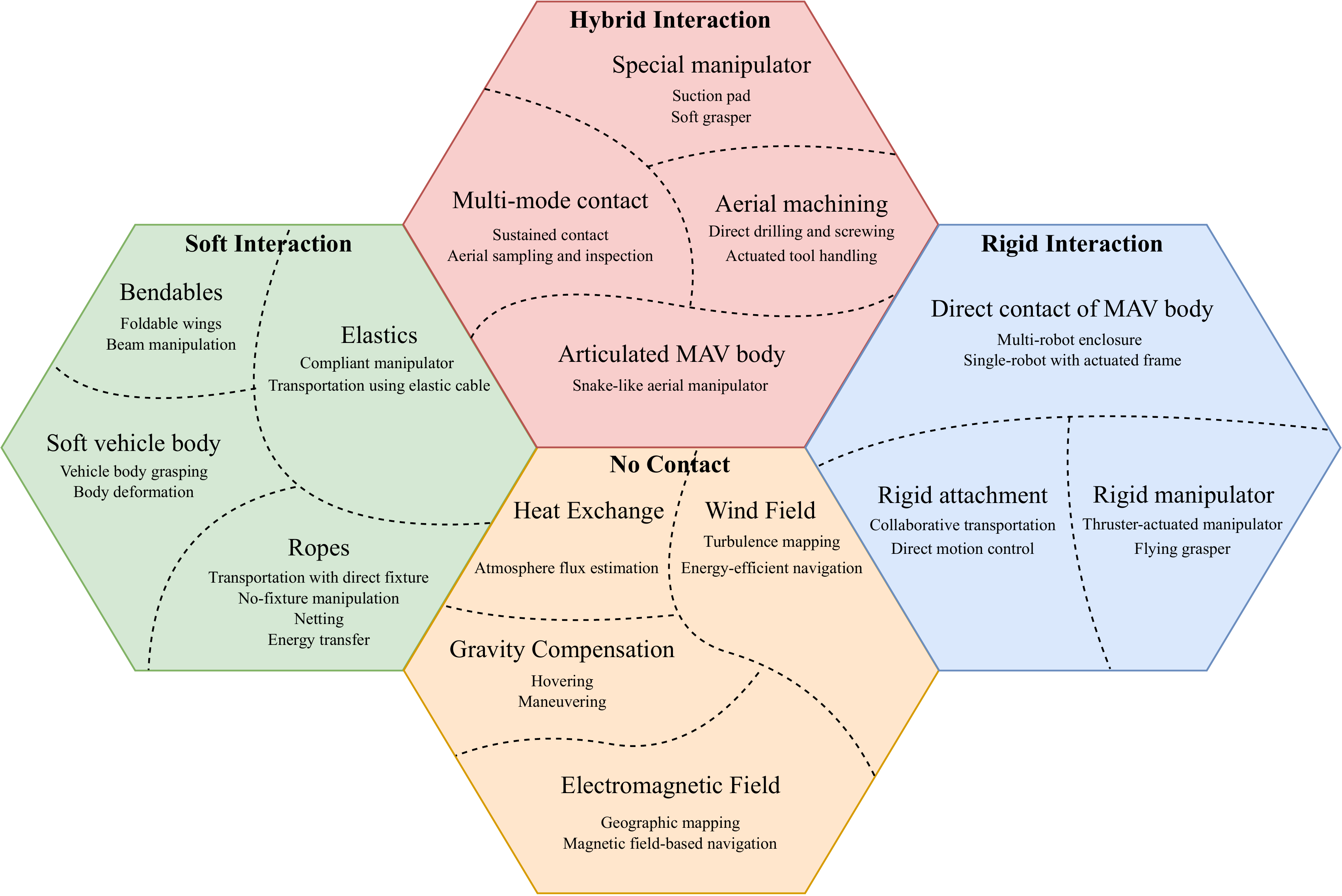}
        \caption{The map of different types of physical interaction enabled by MAVs based on the rigidity of the entities in the model. Each color block represents one type of interaction, and each region of the block represents one mechanism or approach to initiate that type of interaction.}
        \label{fig:mapofphysicalinteraction}
    \end{figure*}
    
    In operation, a MAV often undergoes external and/or internal forces and torques. Simultaneous forces and torques are referred to as wrenches~\cite{10160555}. We generalize these forces and torques as the \emph{influence} received by the MAV. Depending on the task the MAV is tackling, it needs to use or compensate for some of the influence to accomplish the task, which is treated as the \emph{significant influence} in the task. For example, when hovering, a MAV receives gravity as a significant influence. By counteracting the external gravity force applied by the Earth using its propeller thrust force, an MAV can accomplish the task of maintaining hovering. When transporting a rod-shaped object~\cite{pereira2019pose}, the MAV-rod system treats the internal force between the vehicle and the rod as the significant influence that allows the MAV to track the pose of the rod. We refer to the subsystems of concern in the task whose states are affected by the significant influence as \emph{entities}. In the hovering example, the MAV is the only entity. In the rod transportation example, the rod and the MAV are the two entities.

    \begin{definition}[Physical interaction]
        The \emph{physical interaction} that a MAV performs is the comprehensive process where the significant influence in a task changes the states of the entities and the MAV takes action to mitigate or take advantage of the influence to complete the task. 
    \end{definition}

    Our definition indicates that the physical interaction in which a MAV participates is task-specific. A physical interaction sustains as long as the related significant influence continues to apply on the entities. In a hovering task, if the MAV crashed, the physical interaction, where the MAV as the only entity received gravity as the only significant influence, would terminate. 
    
    Researchers and engineers have developed many different mechanisms that allow MAVs to perform physical interaction. These mechanisms, with the intention of attacking different problems, enable various types of physical interaction. Classifying the types of interaction offers insight into the essence of the problems the MAVs intend to solve, and, in return, provides possible improvements of the mechanism or the control strategy for the MAVs. Intuitively, when involved in a physical interaction, the entities often create contact with each other. Whether or not the deformation of the entities caused by the contact is considered or utilized makes a great difference in the approaches to controlling the physical interaction. Therefore, we provide a categorization of physical interaction types based on the rigidity of the entities in the model. 

    \subsection{Rigid Interaction}{
        \begin{figure*}[t]
            \centering
            \subfloat[LASDRA~\cite{8460713}, rigid interaction enabled by connecting rotors to a manipulator.\label{fig:LASDRA}]{
                \includegraphics[height=3.2cm]{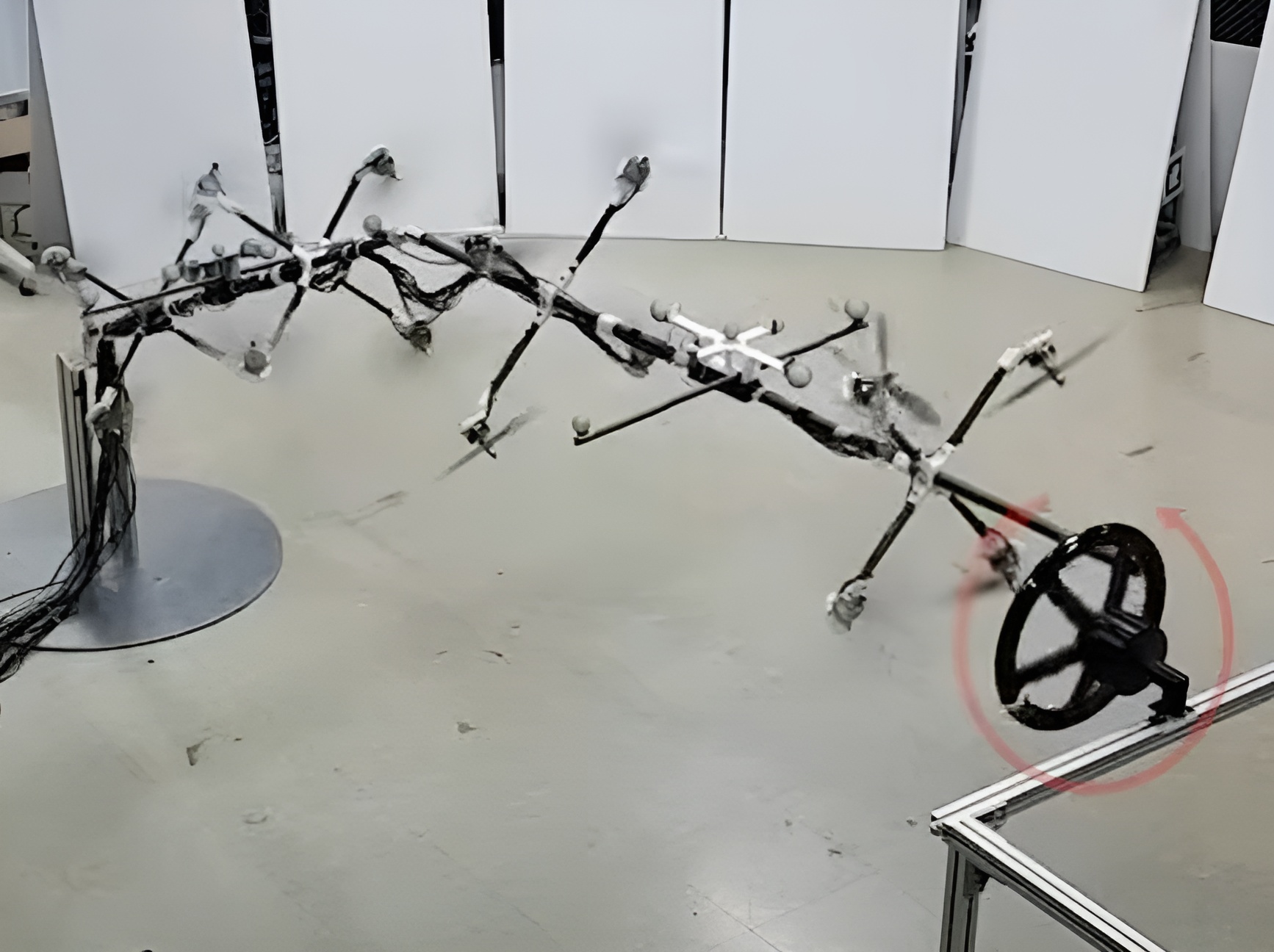}
            }\hspace{0.3cm}
            \subfloat[Collaborative quadrotor transportation~\cite{8120115}, rigid interaction enabled by directly connecting the MAVs to the payload.\label{fig:collaborative}]{
                \includegraphics[height=3.2cm]{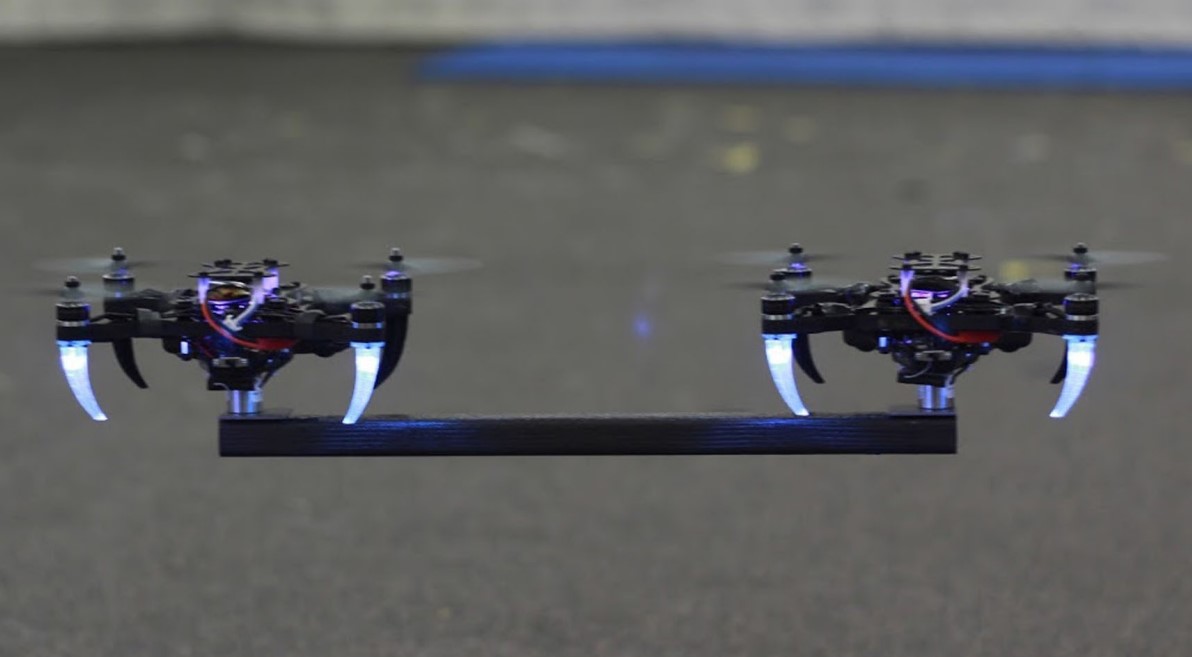}
            }\hspace{0.3cm}
            \subfloat[ModQuad gripper~\cite{8460682}, rigid interaction enabled by enclosing the payload using the frames of the modular MAV.\label{fig:modquad}]{
                \includegraphics[height=3.2cm]{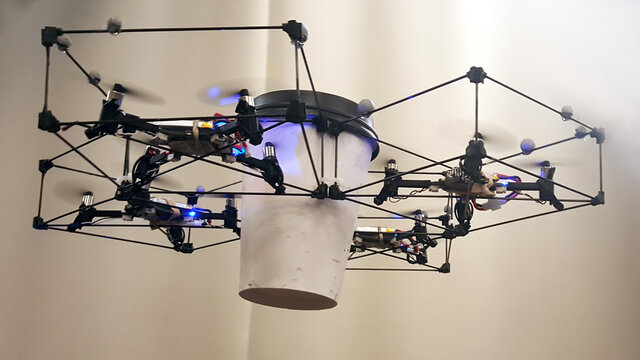}
            }
            \caption{
                Three major approaches taken by researchers to enable rigid physical interaction.
            }
        \end{figure*}
        Rigorous research on rigid body interaction can be traced back to the 16th century due to its universality in the real world and consistent analytical models~\cite{borisov2018rigid}. Such models can effectively describe the bounce of Pingpong balls on the table and, in our case, the trajectory of a flying MAV carrying a payload. Naturally, studies on physical interaction where all entities involved are considered rigid have the longest history and highest maturity among all types of MAV physical interaction. The rigidity in such interaction infers that all entities involved in the interaction are non-deformable, and there exists at least one point on each entity whose relative position to at least one other entity is static, i.e., the connection point. We do not constrain the relative orientation, as a spherical joint connection is allowed. Based on how the rigid interaction is enabled, we introduce three sub-categories.
        
        \subsubsection{Rigid manipulator}
        \label{sec:rigidMAnipulator}
        One of the most starightforward and well-studied approaches that enable a MAV to physically interact with an external object is the rigid connection of a rigid manipulator arm to it, as shown in various flying graspers~\cite{7536029,5980314,7759258,8299552}. In such cases, the entities of the physical interaction are the MAV, the manipulator, and the payload. The rigidity of the manipulator refers to two criteria: the manipulator's links are non-deformable, and its joints are non-deformable unless actuated on command. The task of using MAVs to control the physical interaction can be decoupled into two separate subtasks, i.e., controlling the manipulator and controlling the trajectory of the MAV. 
        
        Pounds et al.~\cite{5980314} deploy a helicopter equipped with a gripper to perform aerial grasping, providing an early example of this approach. The task is divided into three phases: approaching the target, grasping the object, liftoff and departure. Although all three phases are concerned with achieving the hovering trajectory, only the second phase is related to grasping, which creates the physical interaction between the MAV and the payload. Wuthier et al.~\cite{7536029} develop a compact multilink manipulator to be attached to a hexarotor. The authors assume that the flight controller of the hexarotor can compensate for the disturbance caused by the manipulator, and thus focus on the analysis of the arm itself. Mellinger et al.~\cite{6094871} attach a gripper to the bottom of a quadrotor and introduce the disturbance of the gripper payload on the dynamics of the quadrotor as the position change of the center of mass and the additional inertia which are estimated during a period of near-hovering state. This method considers more unknown parameters of the payload, but ignores the continuous motion of the manipulator during the pickup of the payload. On the contrary, Shimahara et al.~\cite{7759258} attach a gripper to the top of a hexarotor, allowing the hexarotor to create torsion along the coaxis of the gripper and the hexarotor. During the interaction, the gripper grasps an object, and the hexarotor controls the torque to rotate the gripper while hovering simultaneously. Orsag et al.~\cite{orsag2014hybrid,8059875} derive a complete dynamic model of a MAV-quadrotor system considering the joint dynamics of the manipulator and the interaction between the manipulator and the quadrotor. The authors further prove the stability of the model reference adaptive control for the system. Such a comprehensive approach ensures stable operation of the MAV even under considerable disturbance due to the interaction.
        
        A different approach under the same category of rigidly combining a manipulator and a MAV is, instead of connecting a manipulator to the MAV, equipping the manipulator with rotors to aid it stay airborne. Boss et al.~\cite{9483143} design a long-reaching aerial manipulator by suspending a bicopter-actuated manipulator with a larger MAV using a long rigid rod. The authors assume that all disturbances are known including those received by the bicopter caused by the manipulation, and treat the entire system as a traditional MAV. The assumption may be too ideal for real-world applications. Yang et al. treat rotors as ``external actuation'' in~\cite{8460713} to increase the strength of an arm-like dexterously-articulated robots called LASDRA, which utilizes the same principle of distributed actuation of MAVs as we mentioned above. Although the LASDRA does not stay airborne and needs a fixed station as shown in Figure~\ref{fig:LASDRA}, the redundant system with many rotors can be directly transferred to a flying multilink manipulator by adding three positional degrees of freedom to its model. Moreover, the only entity in a LASDRA system is the robot itself. The goal of the physical interaction task is to achieve a certain state to prepare for the tasks of a next next, which distinguishes such a system from a pure aerial manipulator. 
        
        \subsubsection{Rigid attachment}
        \label{sec:rigidAttachment}
        Although attaching a manipulator rigidly to a MAV is straightforward, the model can be further simplified by removing the moving components of the manipulator. As a result, entities in the physical interaction are always rigidly connected, leading to less uncertainties in the models. We remark that the rigidity emphasizes on the non-deforming of the entities and connections, instead of that the entire system involved in the interaction must be a rigid body. 
        
        For a different purpose, but with a design similar to LASDRA, Zhao et al.~\cite{8593368} develop a multilink MAV, called DRAGON, that is composed of modular bicopter links connected with spherical joints, which allows the snake-like aerial robot to transform through narrow spatial openings by actively changing the shape of the MAV. The authors apply Linear Quadratic Integral control on the position to maintain the hover of DRAGON, and static feedback linearization to achieve the desired joint angles by applying pseudo-inverse directly on the gimbal dynamics. As discussed in LASDRA in~\ref{sec:rigidMAnipulator}, it is because of the redundancy of the many-rotor system that static feedback linearization in the gimbal control is achievable. Naldi et al.~\cite{7039244} present a ducted-fan based MAV design. By rigidly connecting multiple ducted-fan modules, the MAV is able to increase its actuation capability. The authors apply unified feedback control on the MAV for planar and axial modular configurations in all actuation scenarios. Xu et al.~\cite{9561016} develop a similar MAV design, H-ModQuad, where the MAV modules are independent heterogeneous quadrotors and the controller is derived from~\cite{5717652,5980409}. In all three cases above, we consider each module an entity, and the internal wrenches between modules and gravity are the significant influence. On one hand, DRAGON is different from the other two designs which can be considered rigid bodies because the rigid connection between modules fixes the relative orientation between the modules as well, and thus direct motion control can be applied on the entirety of the MAVs. DRAGON, on the other hand, benefiting from the redundant rotor configurations, also receives direct motion control when its configuration is not singular. 

        \begin{figure}
            \centering
            \includegraphics[width=0.8\linewidth]{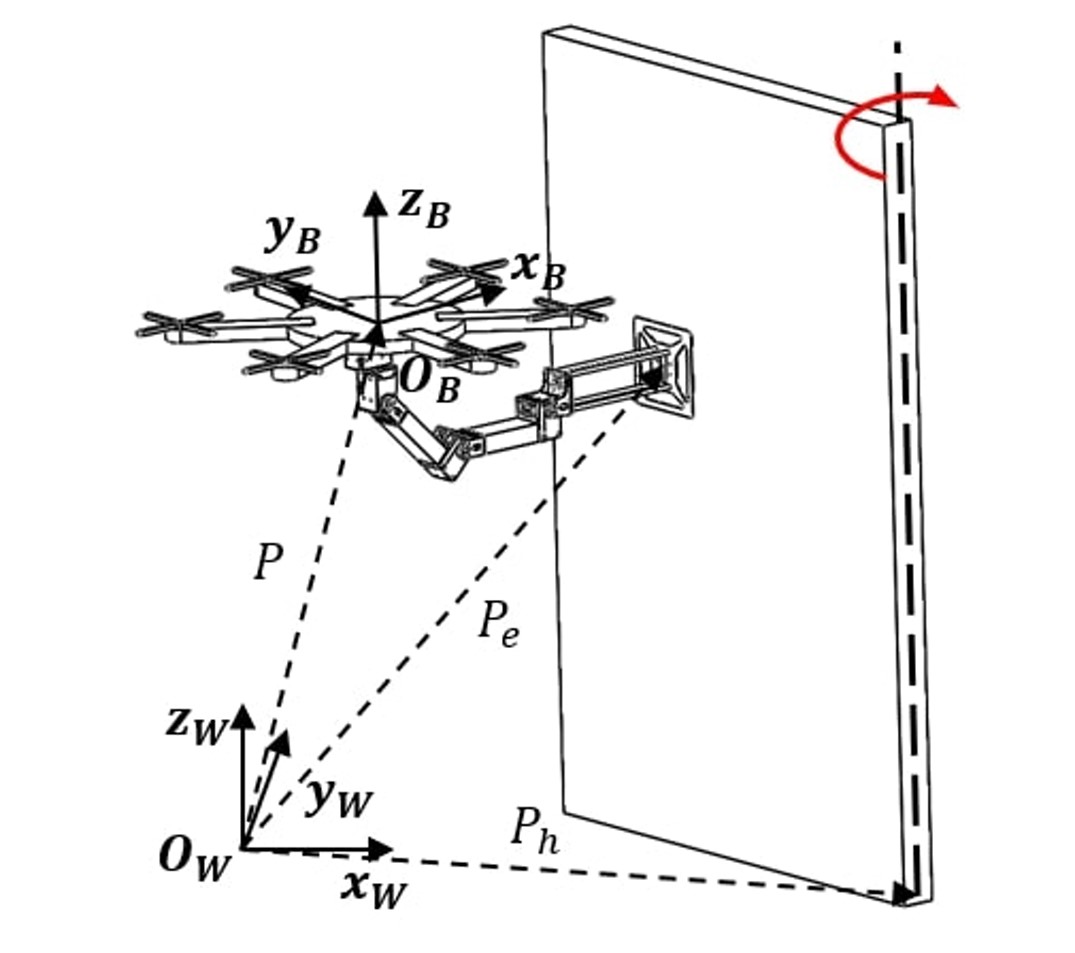}
            \caption{Openning the hinged door using a hexarotor-actuated manipulator rigidly attached to the door~\cite{9197524}: an example of possible cross-category approaches.}
            \label{fig:opendoormpc}
        \end{figure}
        Unlike the attachment of MAVs together, Loianno and Kumar~\cite{8120115} rigidly attach two micro-quadrators to a payload to collaboratively transport it, as shown in~\ref{fig:collaborative}. Similarly to the ducted-fan and H-ModQuad design, the connection between the entities in the physical interaction makes it available to model the entire system as a rigid body. Therefore, a direct motion control can drive it to follow a specified trajectory. To emulate the problem of transporting a manipulator, Pereira and Dimarogonas~\cite{pereira2019pose} connect a rod-shaped object to a quadrotor using a single-axis joint. The task is to drive the quadrotor so that the rod can follow a specified trajectory, where the entities are the quadrotor and the rod, and the significant influence includes gravity and wrenches between the rod and the quadrotor. The authors approach the problem by transforming the dynamics of the MAV-rod system into a form of which the mapping from the input vector to the state vector is known in the literature and a controller is available. Palunko et al.~\cite{6299175} study a similar system in which the quadrotor connects to the payload suspended by the rod with a passive spherical joint. However, unlike~\cite{pereira2019pose}, the rod in~\cite{6299175} is massless. With adaptive control applied, the quadrotor is capable of robustly transporting the load while minimizing swinging behavior, even if the quadrotor is possibly unbalanced. As shown in~\ref{fig:opendoormpc}, Lee et al. combine the method of directly connecting a rigid manipulator arm to a MAV and direct connection between the end effector of the manipulator and an object in~\cite{9197524}. The entities in the interaction include a hexarotor, a multilink manipulator, and a hinged door, all rigidly connected. By applying model predictive control over an objective function of state and input errors, the hexarotor can open the door with minimal energy.~\cite{9197524} is an example where the boundary between a rigid physical interaction enabled by the attachment of a manipulator and that enabled by the direct attachment of a MAV body becomes blurred. Although one can argue that the added manipulator arm can improve the versatility of the system, we might reconsider the design choice of including it since the MAV is rigidly attached to the door without any releasing mechanism. 

        \subsubsection{Direct contact of MAV body}
        A more flexible option to perform rigid interaction is to use direct contact of the MAV body with the other entities, since it does not require any specific mounting mechanisms. Ryll et al.~\cite{7989608} deploy a fully-actuated hexarotor to perform 6D physical interaction with external objects using a passive end-effector extending from the MAV body. The authors apply a geometric controller~\cite{5717652,5980409} to control the wrenches. In addition, through feedback of onboard inertia measurement units and state estimation, the MAV is able to observe the external wrench it receives.
        \begin{figure*}[t]
            \centering
            \subfloat[Catenary robot~\cite{9364354}, soft interaction enabled by ropes.\label{fig:catenary}]{
                \includegraphics[height=4cm]{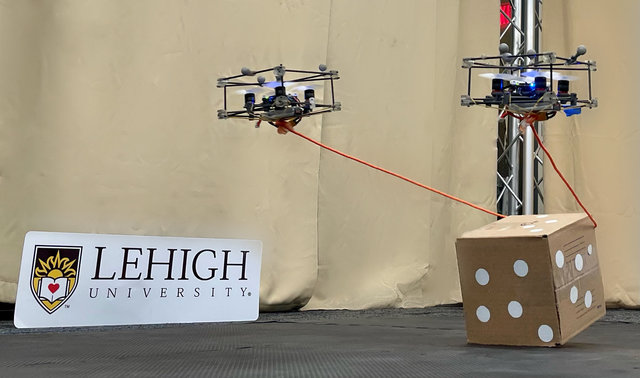}
            }\hspace{0.3cm}
            \subfloat[Octarotor bending a cantilever~\cite{9955376}, soft interaction enabled by bendables.\label{fig:cantilever}]{
                \includegraphics[height=4cm]{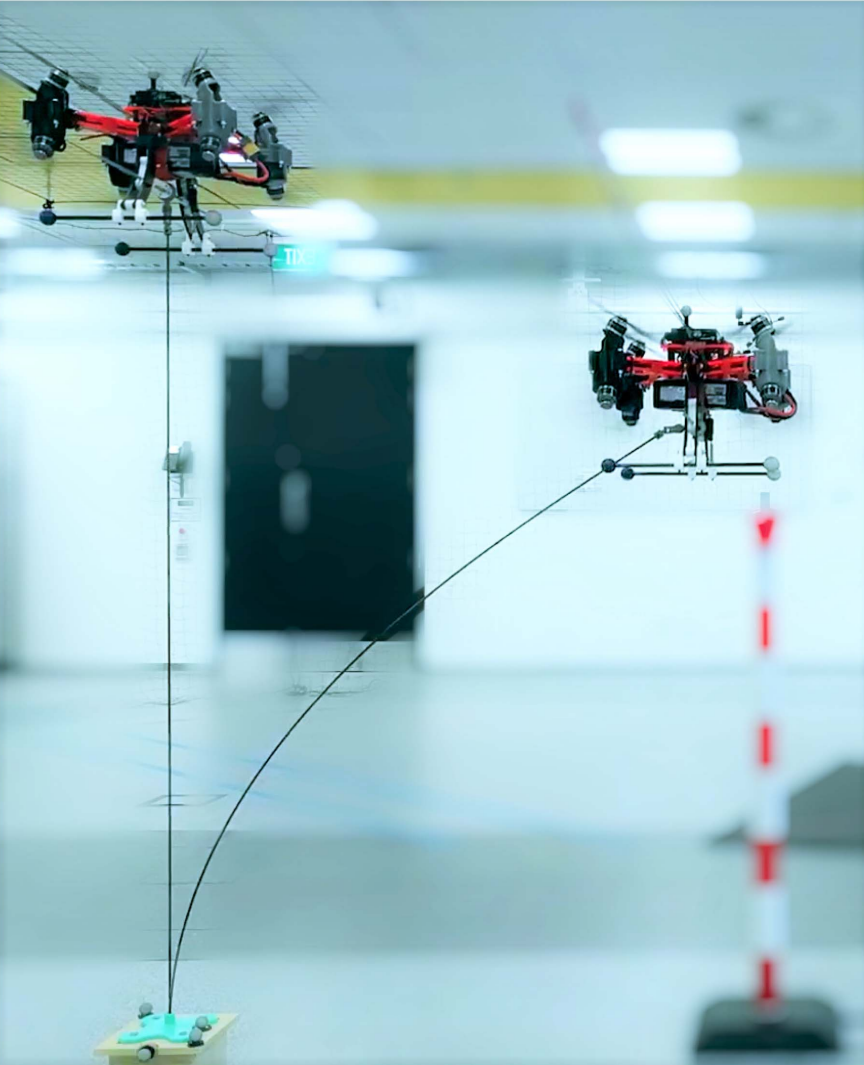}
            }\hspace{0.3cm}
            \subfloat[Aerial vehicle tying hitches~\cite{10160741}, soft interaction enabled by creating enclosure using robot bodies.\label{fig:hitch}]{
                \includegraphics[height=4cm]{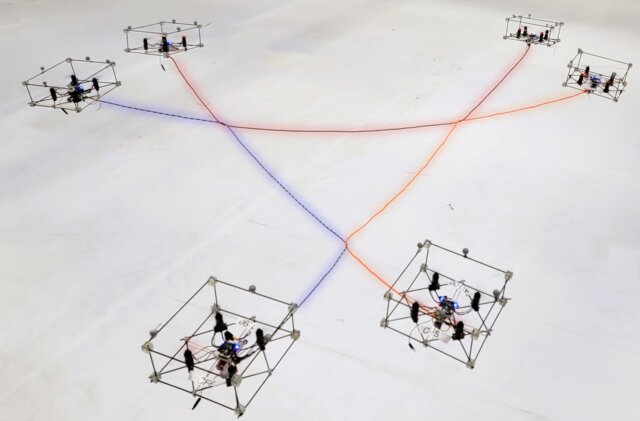}
            }
            \caption{
                Three different mechanisms deployed by researchers to enable soft physical interaction using MAVs.
            }
        \end{figure*}
        Zhao et al. in~\cite{0278364918801639} show the precursor of DRAGON, where each link is a monocopter, and the links are connected via single-axis revolute joints. The multilink MAV is able to create body enclosure on an external geometric object while minimizing joint torque requirements to maintain the enclosure and transport the enclosed object in a similar fashion to multicopters. Gabrich et al.~\cite{8460682} utilize the cuboid frame of ModQuad modules~\cite{saldana2018modquad} and create a flying gripper using only MAV frames by aligning four modules in a $2\times2$ square shape. By driving the diagonal modules to rotate in the same direction and those on the same side in opposite directions, the square is able to articulate and opens the center spaces to rigidly grasp geometric objects. The interaction DRAGON and ModQuad gripper perform both involve the external object and all modules in the MAV as the entities. Thus, the complexity of both systems is high. However, by taking advantage of their redundancy and modularity, both are able to perform aerial manipulation even without a specialized manipulator.
    }

    \subsection{Soft interaction}{
        Although rigid body dynamics can describe a large majority of the physical interaction in the real world with high precision, all macroscopic substances deform in reality. Therefore, when the materials of the entities involved in an interaction deform to a considerable degree, one needs to apply more comprehensive models to describe the interaction more precisely. However, the inclusion of deformation drastically increases the complexity of dynamic models of physical interaction. For example, studies on elastic objects started almost the same time as rigid body dynamics~\cite{euler1980rational} in the form of elastic theory. However, not until the late 19th century did scholars start to formalize general elastic theories into modern mechanics~\cite{gray2001sph,caflisch1984nonlinear}. Furthermore, elastics are not the only types of deformable object that a physical interaction can involve. Soft materials, such as cloth, ropes, and papers, have a long history in human society, but their response to external influence are yet to be fully described in clear mathematics. Although the history of significant research on deformable bodies in the field of robotics is even shorter~\cite{lee2017soft,elango2015review,rus2015design}, researchers find that, by incorporating deformable components, some robotic systems may perform better, especially when the robot intends to deal with uncertainties in a manipulation task. This observation also applies to aerial robotic systems involved in physical interaction such as MAVs. We provide a categorization of soft interactions for MAVs based on the type of soft components that enable them.

        \subsubsection{Ropes}
        \label{sec:rope}
        One of the most popular soft components equipped to MAVs are ropes. Ropes are such long and thin (length $\gg$ radius of cross section) non-rigid entities that do not have a natural shape. They allow none to little extension upon traction and allow none compression. When no traction is applied on a rope, it becomes slack and the internal tension is zero. Since they can be extremely low cost and create tension when receiving traction on two arbitrary points, robotists often use them to create direct non-rigid connection between two entities in an interaction~\cite{8299552}. 
        
        A typical application is shown in~\cite{7536040}, where Pereira et al. use the rope to connect the MAV to a payload point mass, assuming that the rope is massless. Similarly to the authors' other work~\cite{pereira2019pose} mentioned in Section~\ref{sec:rigidAttachment}, the authors apply state-output transformation so that a common quadrotor controller can be directly used to track the pose of the payload. 
        Cruz and Fierro investigate the same system using a different approach in~\cite{6981439}, where the mass of the payload is unknown. During the raise process, the MAV solves a nonlinear optimization problem using the BGF method to estimate the unknown mass. Once the mass is estimated, the MAV can transport the payload using the geometric controller~\cite{5717652}. Palunko et al.~\cite{6631276} improve the tracking performance of the system by applying a least-squares policy iteration to provide an additional deviation term on the position control input from the baseline attitude control. In the above examples, although the ropes are among the entities of the interaction, they are almost always under uncontrolled traction. Thus, they offer little difference from a rigid massless rod connecting the MAV and the payload as long as the tension is non-zero. 
        
        Xiong et al.~\cite{xiong2022optimal} collaboratively use multiple MAVs and unmanned ground vehicles (UGV) to aid a cable-towed manipulator stay airborne. By controlling the cable lengths between the UGV and the manipulator and the pose of the MAVs, the authors implement optimal control to minimize the energy usage of the UAVs while maintaining tension in the cable. Unlike suspended payload transportation tasks, MAVs in~\cite{xiong2022optimal} directly control one of the significant influence in the interaction, the tension in the rope.
        Most ropes with which MAVs interact are considered massless in the robotics literature. However, D'Antonio et al.~\cite{9364354,9571068} point out that when a rope with mass is fixed on two points and under zero external traction, it forms a catenary curve when receiving gravity as a significant influence, as shown in Figure~\ref{fig:catenary}. By connecting the two end points of the rope with MAVs, the system is able to track the lowest position of the curve. When two catenary ropes are coordinated together, they can form knots in the air~\cite{9981363}. When multiple knots are coordinated together, they can form hitches in midair~\cite{10160741} that act like soft grippers as shown in Figure~\ref{fig:hitch}, or a net. Such applications highlight the special physical characteristics of ropes as a soft component in the aerial system. Recognizing the ropes as another important composure of the soft interaction, the group of MAVs are able to use its distributed and flexible actuation to create formulations that further prepare it for interaction with other entities in a new task, for example, aerial manipulation.
        
        \subsubsection{Bendables}
        \label{sec:bendable}
        Considering ropes on the softer side of the components that enable soft interaction, bendables are on the more rigid side among all types of non-rigid components. Bendables are such objects that has a natural shape when under no external influence. They allow little to none extension or compression under external influence. However, they create counter force to the external influence with the intention of recovering to its natural shape. MAVs physically interact with bendables with two main purposes: a) trying to manipulate the bendable to explore approaches to real-world problems; b) allowing the MAV to recover from unexpected external disturbances. 

        Chen et al.~\cite{G005914} deploy two quadrotors to collaboratively transport a bendable rod, referred to as a flexible payload in the paper. The authors consider the payload operating at small bending deformation and use two quadrotors attached to both ends of the rod to transport the rod at near-hovering state. By introducing an integral term in the feedback Proportional-Integral-Derivative (PID) controller, the MAVs are able to contain the elastic potential energy caused by the rod bending in the aerial system without losing stability. Such a containment of elastic energy requires a constant thrust compensation, resulting in potential energy loss during transportation. Given a more comprehensive understanding on the significant influence of the elastic force caused by the bending, one may be able to recycle the energy to make the physical interaction more energy-efficient. Souza and Stol~\cite{9955376} conduct study on how the nonlinear constraint introduced by a bendable rod can affect the performance of a MAV as shown in Figure~\ref{fig:cantilever}. The authors connect a fully actuated octarotor to the cantilever using a rigid rod. The rod connects rigidly to both the MAV and the cantilever, prohibiting both position and orientation changes on both connections. The authors apply Udwadia–Kalaba Formulation~\cite{zhao2018udwadia} based on the Lagrange of the cantilever-rod-octarotor system to obtain its constrained dynamics under high bending deformation. The results of the experiments, where the octarotor tries to control the position of end point of the beam using a traditional PID cascade controller, show the correctness of the model. However, a more effective and dynamic controller is yet to be devised. 

        Instead of actively bending a bendable rod, Alvin et al.~\cite{9812294} choose to utilize the special properties of bendables to mitigate the negative effect of crashing failures of a MAV. The authors attach an array of bendable nylon fiber wings to the MAV frame. During normal operation, the wings are bent and suspended by tension supports on the frame. When a crash occurs, the bendable wings can first partially absorb the impact. During the free fall after the failure, the bendable wings receive wind flow high enough to break the constraint of the tension support and open to their full extent, which is their natural shape. The fully extended wings then start to auto-spin under aerodynamic effects, which significantly reduce the terminal speed of the falling MAV. The whole process, without deriving system identification and parameter estimation, serves the purpose of successfully mitigating an MAV crash by utilizing the physical interaction involving the bendable wings. Arguably, understanding the recovery model of the bendable wings analytically further helps determine the ideal natural shape of the wings for different MAVs.

        \subsubsection{Elastics}
        \label{sec:elastic}
        \begin{figure}
            \centering
            \includegraphics[width=0.8\linewidth]{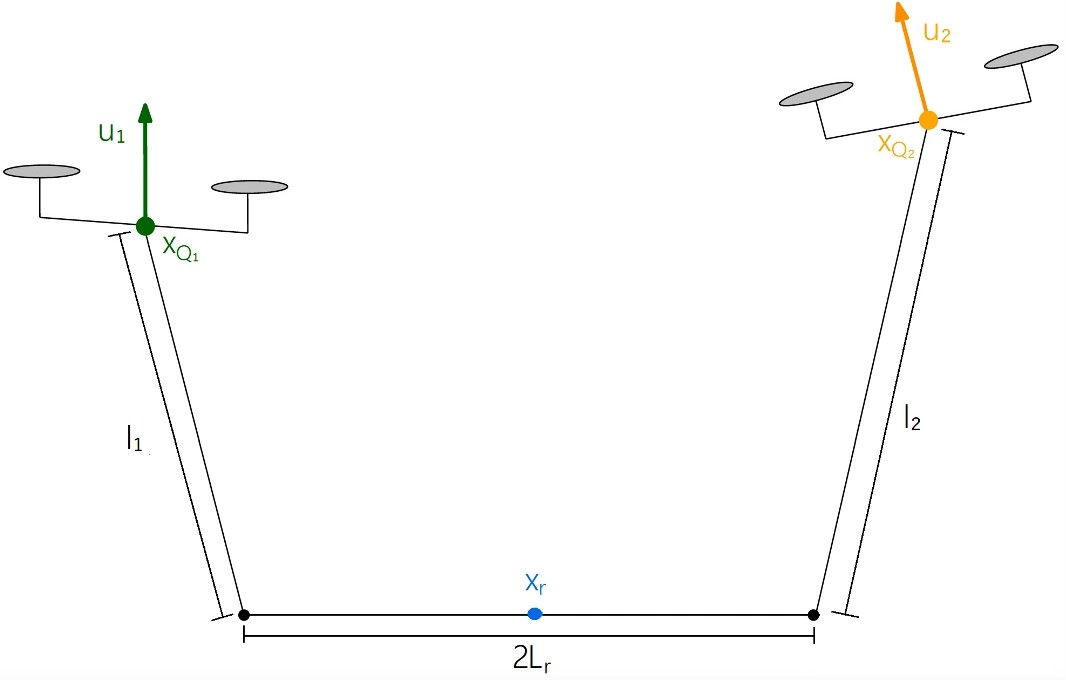}
            \caption{A $2$-D visualization of the elastic-suspended rod transportation problem studied in~\cite{goodman2022geometric}. This case demonstrates how understanding the significant influence in a physical interaction can simplify controller design.}
            \label{fig:elastic}
        \end{figure}
        The ``softness'' of elastics is even higher than that of ropes because they are extensible. Under traction, elastics extend until a breaking point. Some elastics, such as springs, retract under compression, and some others, such as thin elastic bands, may not. The linear deformation model provides a concise mathematical description of elastics with high accuracy, which simply states that the force of the elastic is proportional to its displacement in a given direction. Similar to bendable entities in physical interactions, researchers deploy MAVs to interact with elastics with two purposes: a) validating techniques to control elastics that find applications in real world; b) using the deformation to compensate for unnecessary disturbances. 
        
        Inspired by the problem in tethered payload transportation where ignoring the elasticity of cables may compromise state estimation, Goodman and Colombo~\cite{goodman2022geometric} study the system composed of two quadrotors and one rigid rod payload, as shown in~\ref{fig:elastic}. Each quadrotor is connected to one end of the payload using an elastic cable. By applying singular perturbation theory on the dynamics of the rod-elastic-MAV system, the authors successfully reduce the system dynamics into a similar one with ropes instead of elastic cables, which also reveals the differential flatness of the system. As the pose errors of the rod propagate back the dynamic model of the system to the pose errors of the quadrotors, applying geometric controllers~\cite{5980409,5717652} on the two quadrotors realizes the trajectory tracking of the rigid rod. This approach studies the significant influence between the elastics and other entities. Then, through variable manipulation, the elastic-related terms are reduced from the dynamics of the system. In return, the authors are able to directly apply mature control techniques available in the literature to achieve stable object transportation. 
        
        Researchers often model transportation tasks as payload tracking problems where the payload is suspended by cables or rods. In~\cite{9312497}, Yiğit et al. create a system where an omnidirectional MAV is suspended by an elastic spring on a robotic carrier. Assuming that the displacement of the spring from its free state is known, the authors are able to directly apply PID control and feedback linearization on the MAV to drive it to any desired orientation. The simplicity of the model makes this interaction straightforward to control, since it only contains the MAV as an entity and the elastic can be simplified to the significant influence of the elastic force, similar to gravity. However, the motivation of this project is to enhance existing manipulator designs using MAVs. Showing how the MAV can compensate for unknown disturbances caused by the robotic support may be the direction of improving this work. Fishman and Carlone~\cite{9438502} attach a tendon-actuated soft gripper on a quadrotor. The authors model the problem of controlling the quadrotor-gripper system to maneuver and to grasp objects as a the tasks of tracking the trajectory of both the gripper tips and the quadrotor. By separating the control of the quadrotor and the gripper, the authors apply optimal control on the gripper for aggressive grasping and geometric control on the quadrotor to compensate for disturbance caused by the load. The objective of optimal control includes minimizing the maximum rate of tendon motion, minimizing the distance from the tip of the gripper to the centroid of the object, and maximizing the angle of the enclosure and the distance of the gripper on the object. The inclusion of a tendon-based elastic manipulator enables adaptation to deviations from the nominal quadrotor trajectory and naturally mitigates the impact of contact forces. Similar systems that combine compliant manipulators with MAVs are presented by Khanmirza et al.~\cite{khanmirza2018underactuated}, where the authors implement a path planning algorithm considering the nonholonomic constraints caused by the compliant manipulator.

        \subsubsection{Soft vehicle body}
        \label{sec:softvehiclebody}
        Similar to rigid interactions, MAVs can leverage its body softness to create soft interaction. The main advantage of such MAVs is their versatility: without additional equipment mounted, they can initiate soft interaction any time on command or just stay flying without worrying about the dynamic disturbance caused by transporting the equipment.
        
        Many snake-like aerial vehicles can create body enclosures. Considering the categorization provided by Mendoza-Mendoza et al.~\cite{8979344}, the catenary robots~\cite{9364354} are considered soft robot snakes. When forming hitches around an object~\cite{10160741}, the enclosure created by the cables allows the robot to transport it. As mentioned above, the physical interactions initiated by MAVs do not always involve an external object. Ruiz et al. construct a quadrotor with tendon-actuated rotor arms called SOPHIE~\cite{9851515}. By tightening the tendons, the rotor arms bend. The authors deploy sensors to monitor the angles of bending so that the feedback controller can adapt to the new rotor actuation distribution, where the dynamics of the bending is not considered. Therefore, a linearization-based method is directly applied to control the motion of the quadrotor. Considering the bending a dynamic process, Bennaceur and Azouz design a novel quadrotor with flexible rotor arms~\cite{bennaceur2022modelling} to study the dynamics and controller of quadrotor when the deformation of the rotor arms are non-negligible. Through Lagrange analysis and decoupling the system into two subsystems using the substitution of virtual input, the authors find that one of the two subsystems is flat. They first implement a stabilization controller for the flat subsystem. Then, they apply a backstepping controller on the linearized subsystem that describes the vibration of the rotor arms. 
    }

    \subsection{No contact}{
        Most of the time, establishing contact is necessary for MAVs to perform physical interaction. However, when there is no visible contact, MAVs can interact with the environment as well with the help of special equipment. Due to their flexible deployment, MAVs are suitable for such field applications. For example, equipped with a thermometer, a MAV is able to traverse a large area of atmosphere to estimate the heat flux~\cite{atmos10070363}. The developing technology of wireless energy transfer through electromagnetic fields~\cite{electronics9030461} may allow MAVs to obtain a longer duration of operation. Due to the agility of MAVs, once equipped with a hall sensor, they are able to navigate in the magnetic field of two parallel transmission lines~\cite{app11083323}. Even without a specialized sensor, MAVs are also able to sense wind fields through state estimation~\cite{gonzalez2019sensing}, apply adaptive control to navigate in wind fields~\cite{7152250}, and take data-driven approaches to navigate in wind fields~\cite{wang2021data}. In this survey, we will not focus on such interactions.
    }

    \subsection{Hybrid interaction}{
        \begin{figure}[t!]
            \centering
            \includegraphics[width=0.5\linewidth]{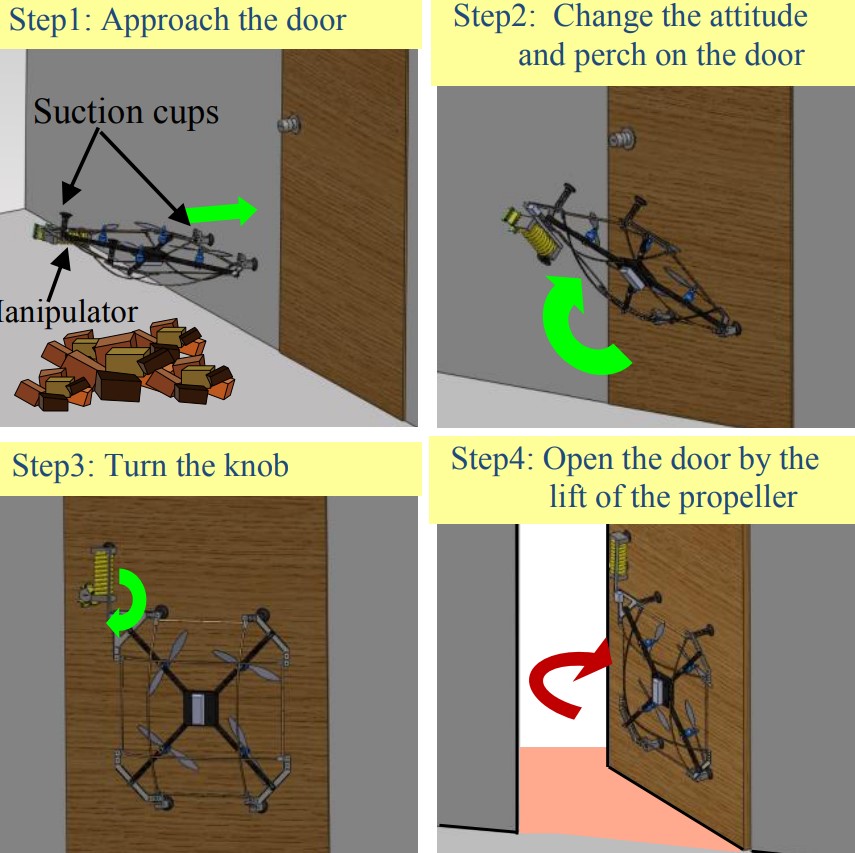}
            \caption{Suction pad perching and manipulation~\cite{7139845}, hybrid interaction enabled by special manipulators.}
            \label{fig:perch}
        \end{figure}
        
        \begin{figure*}[t]
            \centering
            \subfloat[Pogodrone~\cite{zhu2022pogodrone}, hybrid interaction enabled by multi-mode contact.\label{fig:pogodrone}]{
                \includegraphics[height=2.8cm]{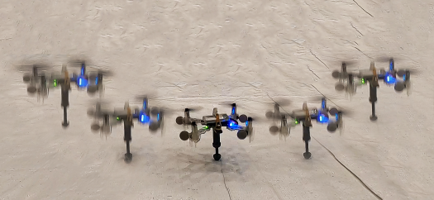}
            }\hspace{0.3cm}
            \subfloat[DRAGON~\cite{zhao2022versatile}, hybrid interaction enabled by articulated robot bodies.\label{fig:dragon}]{
                \includegraphics[height=2.8cm]{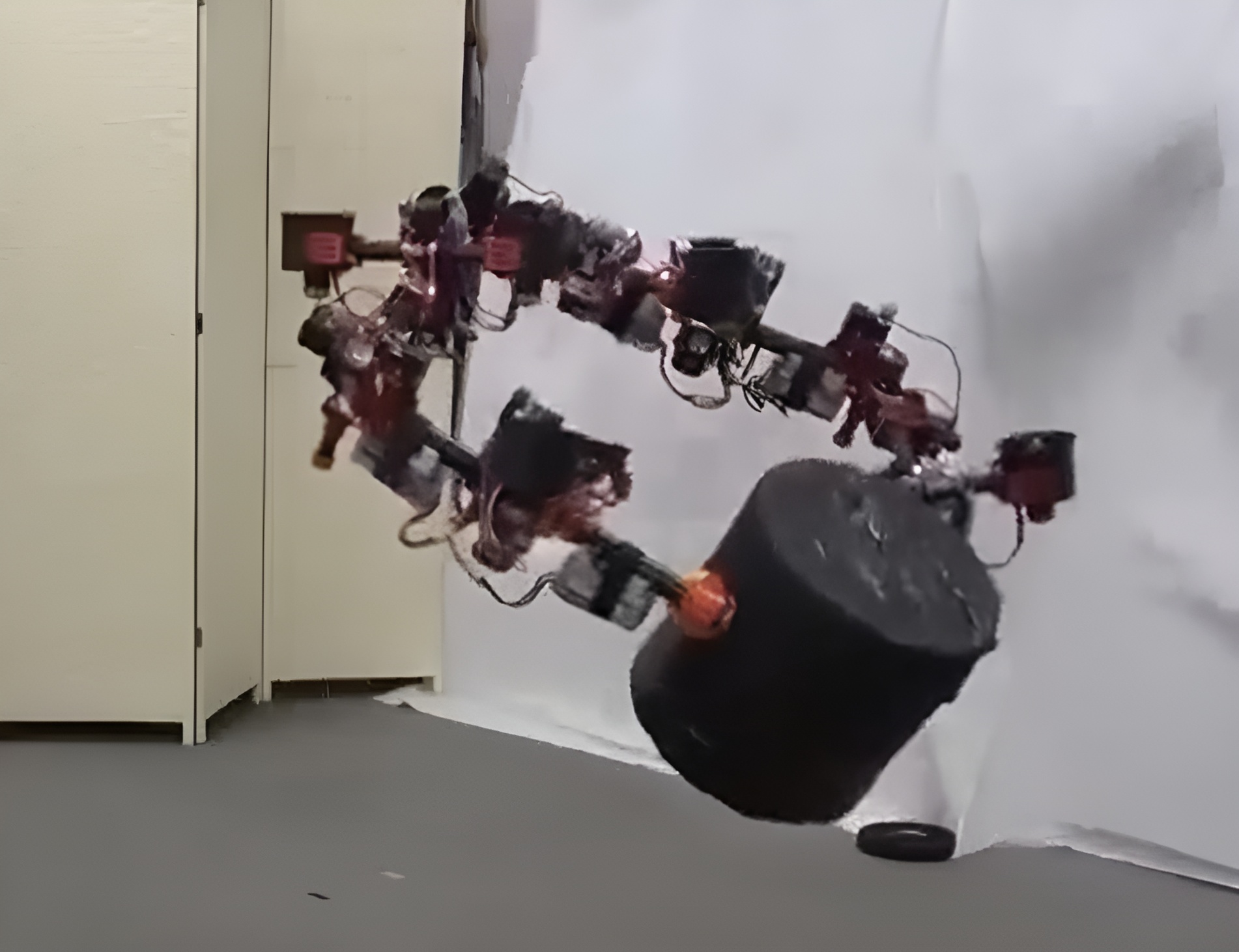}
            }\hspace{0.3cm}
            \subfloat[Aerial drilling on a quadrotor~\cite{9363592}, hybrid interaction enabled by aerial machining.\label{fig:machine}]{
                \includegraphics[height=2.8cm]{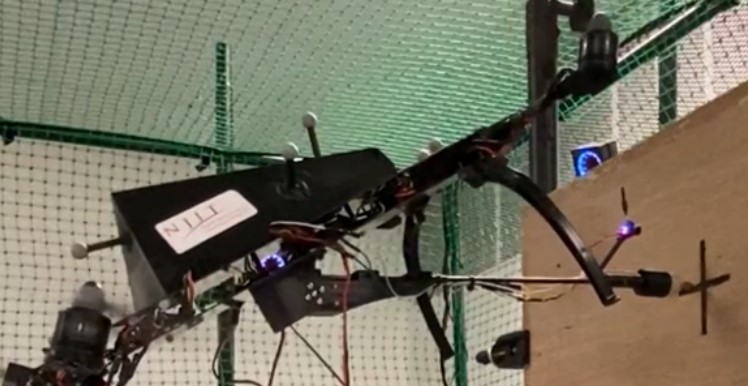}
            }
            \caption{
                Three different approaches of hybrid physical interaction performed by MAVs.
            }
        \end{figure*}
        Some interaction processes contain multiple phases, where the same entities in the interaction experience different types of significant influence. However, forcing the separate discussion on such interactions may lead us to analyzing an incomplete piece and losing the greater picture of the entire physical interaction. We categorize them as hybrid interactions. A significant criterion for identifying a hybrid interaction is to find different phases, where the rigidity changes or the contact breaks between two entities. 

        \subsubsection{Special manipulators}
        MAVs equipped with some special manipulators, such as a suction pad, would initialize the interaction by contacting the manipulated object (soft) to the completion of grasping the object (rigid). Tsukagoshi et al.~\cite{7139845} develop a quadrotor mounted with a suction pad that allows it to perch on a door, and a grasper that turns the door knob. The whole interaction is divided into four steps, each of which performs a unique interaction. Inspired by eagle claws, Tscholl et al.~\cite{tscholl2023flying} design a flying soft-electrostatic claw that is hydraulically amplified to achieve rapid manipulation with the environment through both electrostatic and gripping force. 

        \subsubsection{Multi-mode contact}
        Sometimes, a phase shift may be covert during a hybrid interaction. For example, when using a MAV equipped with a compliant manipulator to maintain contact with a surface~\cite{7387723}, the manipulator may deform on impact with the wall, causing a short period of soft interaction. Once the MAV stabilizes, simply maintaining contact does not deform the manipulator, turning the model into a rigid interaction. Similarly, Bodie et al.~\cite{bodie2019omnidirectional} connect a contact-based inspection tool to an omnidirectional MAV. By applying a model-based selective impedance control, the MAV-inspector system is able to slide on a surface without losing stability. The pogodrone developed by Zhu et al.~\cite{zhu2022pogodrone}, in contrast, offers a clear shift between different phases: when hovering, the MAV is a rigid model; when impacting the ground and taking a soil sample, the pogodstick acts as an elastic spring that turns the interaction soft.

        \subsubsection{Aerial machining}
        Multiple phases of interaction may shift rapidly, especially in aerial machining tasks. Equipping a quadrotor with rotary tools~\cite{9363592} with a 1-DOF manipulator, Ding et al. allow the MAV to actively drill and screw on a wall. The authors apply a robust adaptive control to compensate for unknown disturbance during machining. The interaction involves the manipulator, the tool, the quadrotor, and the wall. In~\cite{5513152}, the quadrotor is connected to a brush. It is also connected with the wall through a hinged support beam. By actuating the rotors, the quadrotor-brush system is able to brush the wall automatically.

        \subsubsection{Articulated MAV body}
        As we mentioned before, the modular DRAGON~\cite{zhao2022versatile} is able to shift between multiple types of interaction on command. The modular nature allows such an articulated system to apply many types of significant influence. In~\cite{zhao2022versatile}, the authors show that the same MAV can hover, manipulate an external object, and grasp with the MAV body.
    }

    \subsection{Discussion}{
        \subsubsection{Understanding the interaction}
        In our categorization, the type of physical interaction depends mainly on the special mechanisms mounted on the MAV and the entities with which the MAV interacts. However, all interactions are initiated and controlled by the MAVs. The methods researchers apply either to compensate for an influence or to utilize it, all need to be converted into rotor actuation to realize. Therefore, we focus our discussion on the actuation of MAVs and consider only the significant influences applied on the MAV by all other entities.
        
        \paragraph{MAV} At the highest abstraction level, we can consider the effect of mounting special tools to MAVs as converting MAV actuation into influences in specific directions or on specific entities. Thus, the ability to control a physical interaction eventually depends on the actuation capabilities of the MAV. Such an abstraction conforms to our observation on the case study. Researchers tend to deploy larger MAVs and/or MAVs with higher actuation capabilities in tasks with higher complexity by connecting more powerful mechanisms. However, we doubt that such a methodology would be practical in the near future if we were to push the research frontier of physical interaction with MAVs. In our opinion, the most essential characteristics MAVs have that makes them stand out among various robotic systems is their nature of distributed actuation. In other words, the fact that one can add the thrust force generated by each rotor of a quadrotor to obtain its total force makes them scalable robots by design. Therefore, instead of trying to strengthen individual MAVs, further exploration of the collaborative behaviors of MAVs in physical interactions will be the direction of future research. Moreover, modular MAVs can change their actuation capabilities on command depending on their module configurations. Their redundancy not only makes them more resilient against failures but also allows them to tackle different tasks without much hardware modifications. Therefore, we would like to conclude our discussion on MAVs involved in physical interactions as such: collaborative physical interactions carried out by modular MAVs will edge future research on MAVs.
        
        \paragraph{Questions around significant influence}
        When dealing with the significant influence in physical interactions, the researchers typically let the MAVs generate actuation to control the interaction. However, as shown in~\cite{9812294}, the MAV rotors were assumed to stop working. Simply by passively interacting with airflow and let the bent wings release their stored energy, the MAV can safely land. Such an application shows a possible direction of treating the significant influence of an interaction. The question that we are unable to answer in this survey is whether there exists a universal approach to passively utilizing the significant influence in a physical interaction using a MAV. Furthermore, in~\cite{goodman2022geometric}, the authors show that the elastic force in the cables that connect two quadrotors to the payload is effectively insignificant. It raises another question that we are unable to answer in the survey, which is whether we can find a universal approach to ignore the significant influence of a physical interaction. Only by digging into mathematical models of the interactions can we figure out the answers. 
        
        \subsubsection{Passages}
        In the discussions, we notice some MAVs are able to participate in different types of interactions such as the DRAGON~\cite{zhao2022versatile,8593368,0278364918801639} and ModQuad~\cite{saldana2018modquad,8460682,xu2022modular}. Based on the current task, the same MAV is able to transition from one type of interaction to another. Although we prepare the map of physical interactions shown in Figure~\ref{fig:mapofphysicalinteraction} with separate blocks and regions, there exist passages from one region to another in the same block or even crossing blocks. Some transitions are achievable with similar MAVs under different command. Some may require modification of the physical components. For example, we categorize the manipulation initialized by MAVs equipped with the suction pad~\cite{7139845} under hybrid interaction because there exists a phase shift. However, if the suction pad is replaced with another grasper, the interaction becomes rigid.
        Even two distinct MAVs performing different types of interaction could transform from one to another, such as LASDRA~\cite{8460713} and DRAGON by changing the actuation and the control policy.
    }
}

\begin{figure*}
    \centering
    \includegraphics[height=10cm]{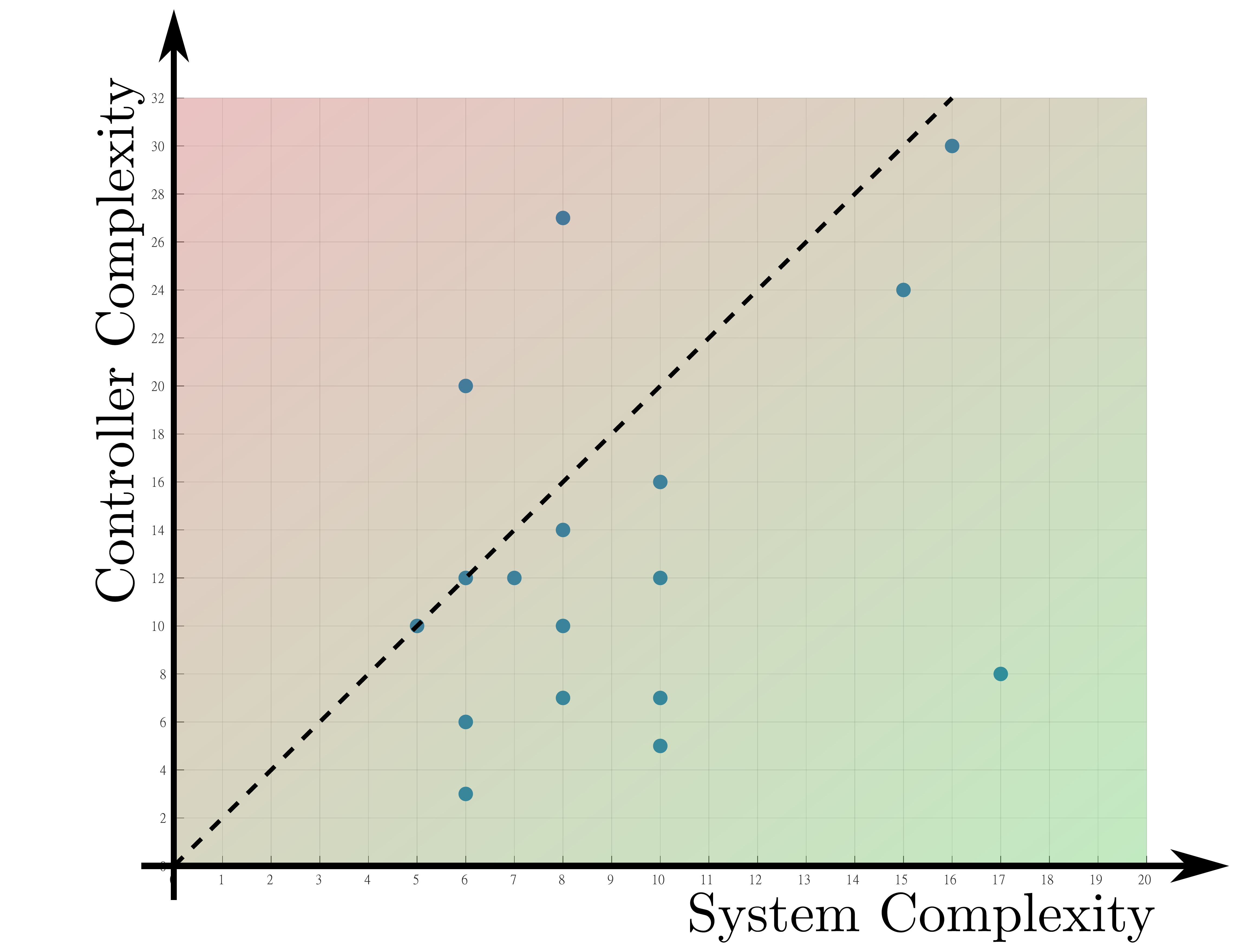}
    \caption{The evaluation plot of the specified controllers applied on physical interaction systems studied in this survey. The horizontal axis reveals the system complexity and the vertical axis represents the controller complexity, both defined in the beginning of~\ref{sec:control}.}
    \label{fig:evacon}
\end{figure*}
\section{Control Strategy}{
    \label{sec:control}
    When performing physical interaction, MAVs often encounter uncertainties. For example, when making contact with a surface in the environment, the MAV does not always know the model of the contact, let alone the model parameters~\cite{6875943}. To deal with these unknown factors, MAVs must adjust to unexpected forces or torques during physical interaction using feedback control. Based on the types of interaction, different control strategies are more applicable and effective. We would like to evaluate four types of controllers seen in the literature deployed on MAVs to perform physical interaction, and compare the complexity of the controllers applied in different physical interaction tasks. More specifically, we quantize the complexity of the task by the dimension of the generalized coordinate of the models~\cite{7759268,corke2023mobile}, and the complexity of the controller by the number of control parameters. In the case of optimal control, we add the maximum dimension of the constraints present by the optimization to the complexity of controllers. For example, in~\cite{5717652,5980409}, a quadrotor in $3$-D space has $3$ position coordinates and $3$ orientation coordinates, summing up to $6$-D generalized coordinates. The geometric controller that controls the quadrotor requires proportional and derivative (PD) gains for the errors of each generalized coordinate, resulting in $12$ control parameters. gains Note that our metric of controller complexity excludes the difficulty of \emph{deriving} the controller, but provides an estimate on the overhead of onboard computation since MAV attitude control is latency sensitive~\cite{9811564}. We admit that these metrics do not reflect the complexity caused by the nonlinearity in the model or the controller, and are not suitable for controllers derived in frequency domain. However, it shows how 
    
    \subsection{PID and linearization}
    The most common control strategy for controlling MAVs during their physical interactions is to apply PID feedback on state errors to obtain the desired actuation. PID stands for Proportional, Integral, and Derivative; each represents a gain coefficient that multiplies with a specific error term. The summation of all the multiplications gives the desired actuation in the generalized coordinates, respectively, to the state. Almost all cases we studied in this survey apply some form of PID feedback.

    Obtaining desired actuation is only the first step of controlling a MAV.
    One of the biggest advantages of MAVs is that most of them are differentially flat, which means that we can always find a set of outputs such that all states and inputs can be determined from these outputs without integration~\cite{9794477}. Researchers use this property to apply feedback linearization to eliminate the nonlinearity in the systems so that the desired input can be obtained through algebraic operations between the observed states and the desired actuation. The advantages of feedback linearization include high computational efficiency, ease to derive, and high responsiveness. Whenever a MAV allows feedback linearization, the developer should choose it, as shown in our case studies~\cite{goodman2022geometric,7989608,8120115,7039244,9561016,6981439,6631276,9312497,9364354}.
    
    However, when performing physical interactions, the dynamics of MAVs may not accept feedback linearization, or some required feedback is not available, as in~\cite{bennaceur2022modelling}. One method of overcoming this issue is trying to decouple a subsystem from the full system, then attacking other subsystems that may already be of less complexity due to the decoupling. Another method is choosing one equilibrium of the system and linearizing the system around the equilibrium. In operation, the gains are carefully chosen to ensure that the system operates near equilibrium~\cite{0278364918801639,8593368,9851515,6094871}. By applying Lyapunov analysis~\cite{sastry1999lyapunov}, sliding mode analysis~\cite{shtessel2014sliding}, pole placement~\cite{6299175}, or other nonlinear control techniques, one can also devise nonlinear controllers to stabilize the MAV under the influence of entities. 
    
    \subsection{Adaptive control}
    Applying adaptive control does not exclude other control techniques from being applied to the system. It is typically introduced into a system to compensate for the uncertainty of parameters, such as the center of mass during transportation~\cite{6299175} or the contact point between a cable and a manipulated box~\cite{9571068}. By separating the adaptation term that contains the unknown parameters of a system, one can rewrite the error dynamics of the system in terms of a regressor and the unknown parameters. By updating the unknown parameters using the errors, the baseline controller, which typically contains the unknown parameters, is able to drive the MAV with increased performance. 

    Adaptive control is suitable for adjusting parameters of a known model that would not cause fatal failures when incorrect because it requires iterative updates based on system errors to reach a reasonable estimation. For the same reason, parameters that change rapidly may not be suitable for an adaptive control to estimate.
    
    \subsection{Optimal control}
    Optimal control is becoming more prevalent in recent literature, as the onboard computation power of MAVs has increased dramatically due to hardware advancement. In~\cite{9438502}, the authors model the entire aerial manipulation task as one optimization problem. In~\cite{9197524}, the authors control the door opening behavior using the model predictive control to achieve minimal energy usage. In the early work on the DRAGON~\cite{8593368}, the authors apply Linear Quadratic Integral control to track the hovering position of the MAV, which is another type of optimal control. Compared to other control techniques, optimal control for MAVs in physical interaction allows the inclusion of actuation constraints and an objective function, ensuring that the generated control actions satisfy the specified limits. However, optimal control requires an accurate model. If the physical interaction possesses an unmodeled disturbance, the MAV may fail the task trying to perform the interaction.The computational complexity is another problem of optimal control, especially when the dynamics of the physical interaction is highly nonlinear. The latency caused by solving optimization problems may be detrimental to sensitive systems. 
    
    \subsection{Learning based control}
    In highly unstructured systems, one may want to apply learning-based control techniques. However, very few applications where MAVs perform physical interaction are found in the literature using learning-based controllers. In this survey, we mention two, one trying to improve the performance of cable-suspended transportation using least-squares policy iteration~\cite{6631276}, another trying to navigate the MAV in a wind field. Although both achieve the control goals, both do not exhibit significant improvements in comparison to other model-based solutions. Learning-based controllers typically require training data, online or offline. Both demand prior successful deployments of the MAVs in the interaction. The controller based on the trained policy/neural network is also much more computationally expensive, which keeps researchers from applying on real embedded systems on which controllers of MAVs typically operate. Even the most basic policy network contain dozens of nodes, resulting in a high number of parameters. Therefore, we exclude them from the complexity evaluation.

    \subsection{Complexity evaluation insights}
    We plot the comparison of the complexity of the controller system in Figure~\ref{fig:evacon}. The dashed line marks the baseline where the complexity of the controller is twice the complexity of the system, the same as the geometric controller~\cite{5717652,5980409}. Each blue dot represents a case studied. We notice that most applications are below the baseline. If we are able to use the smallest number of control parameters to stabilize a system in a high-dimensional generalized coordinates, it means that either the system is constrained and it loses degrees of freedom, or the system possesses natural stability. In our case, the blue dot in the lower right corner belongs to~\cite{goodman2022geometric} where the authors model all entities in the interaction as free bodies and then mathematically reduce it considering all constraints. In other words, the closer an approach is to the lower right corner of the plot, the further it simplifies the model of the interaction. 

    In contrast, if an approach is closer to the upper left corner of the plot, it may infer that the controller is over-constraining the actuation of the MAVs in the interaction, and that the controller may not be widely applicable to other types of interaction. 

    Another observation that is worth mentioning from the plot is that if an approach is closer to the upper right corner, it indicates that the system is highly actuated and is able to perform many different interactions as the controller is constraining the actuation under a fair number of requirements.
}

\section{Conclusion}{
    In this survey, to support our answer to the question on the purpose of multi-rotor aerial vehicles, we conducted a thorough review of the literature on multi-rotor aerial vehicles involved in physical interactions. We defined physical interactions and provided a nomenclature system of subjects involved in the interaction. We categorized the interactions into four groups based on the rigidity of interaction models and analyzed each approach we studied using the notation of physical interaction. For some of the approaches in the literature, we highlighted their special attributes and shared our opinion of possible improvements. We concluded that, for future research on MAVs, we would explore the area of collaborative physical interaction using modular MAVs. We also had a shorter, specific discussion on the control strategies applied to control the physical interactions in the case study. We provided a quantitative evaluation of the complexity of the controller and provide intuitive insights from the evaluation plot. 
}

\section{My Progress}{
    I would like to highlight my progress in the doctoral program at Lehigh University so far, considering my future focus on the area of collaborative physical interaction using modular MAVs. In~\cite{9561016,xu2022modular}, we designed a modular multi-rotor aerial vehicle with heterogeneous modules, which can be the base platform for our future experiments and applications because it can change the actuated DOF based on the configuration. The design is further formalized in~\cite{dcustomizable}. In~\cite{zhu2022pogodrone}, we designed the PogoDrone, where we implemented a hybrid controller that allowed the robot to transition between different types of interaction. It provided me with a baseline for meaningful physical interaction using aerial vehicles. In~\cite{cui2023toward}, we deployed a low cost tactile sensor on a manipulator arm and successfully achieved control of the contact mode using a learning-based approach. This work provided me with a baseline of meaningful physical interactions in the manipulation field. In~\cite{10160555}, we developed a theoretical framework to analyze the ability of the MAV to activate and a search algorithm to find a modular MAV configuration, given the desired task requirements. An upcoming publication on a quadrotor-actuated blimp system opens another platform to explore physical interaction with. My ongoing and future projects include dynamically controlling a system composed of two quadrotors connected by a bendable object, optimal bicopter blimp control, imitation learning in mobile vehicles, and collaborative aerial sawing. The scope of these projects is well aligned with this depth study.
}

\bibliographystyle{IEEEtran}
\bibliography{ref}

\end{document}